\documentclass{article}

\usepackage[preprint]{neurips_2025}
\usepackage{amsmath,amssymb,amsfonts}
\usepackage{graphicx}
\usepackage{textcomp}
\usepackage{multirow}
\usepackage{xcolor}
\usepackage{braket}
\usepackage{algorithm}
\usepackage{algorithmicx}
\usepackage{algpseudocode}
\usepackage{graphicx}
\usepackage{booktabs}
\usepackage{subcaption}
\usepackage{quantikz}

\usepackage[utf8]{inputenc} 
\usepackage[T1]{fontenc}    
\usepackage{hyperref}       
\usepackage{url}            
\usepackage{booktabs}       
\usepackage{amsfonts}       
\usepackage{nicefrac}       
\usepackage{microtype}      
\usepackage{xcolor}         
\usepackage{comment}

\title{QeHDC: Hyperdimensional Computing based on Quantum-enhanced binding and SuperClass Construction}

\author{
  Yangjie XU\\
  University of Luxembourg\\
  Kirchberg, Luxembourg\\
  \texttt{yangjie.xu@uni.lu} \\
  \And
  Hui Huang \\
  University of Luxembourg\\
  Kirchberg, Luxembourg \\
  \texttt{hui.huang@uni.lu} \\
\AND
  Li Ning \\
  Stellaris AI Limited \\
 Hong Kong, China \\
  \texttt{lining@stellaris-ai.com} \\
  \And
  Radu State \\
  University of Luxembourg\\
  Kirchberg, Luxembourg \\
  \texttt{radu.state@uni} \\
}

\begin{document}

\maketitle

\begin{abstract}
Hyperdimensional Computing (HDC) is a robust computational framework inspired by human cognition characterized by simple and efficient operations within high-dimensional vector spaces. Quantum-enhanced Hyperdimensional Computing (QeHDC) extends classical HDC by leveraging quantum mechanical properties to enhance computational efficiency. In this paper, we propose a novel Quantum HDC framework featuring a one-pass training method, leveraging sinusoidal and quantum encoding to project classical data into quantum amplitude states efficiently. Our framework introduces an innovative reference-state-based quantum binding operation realized via quantum circuits. Furthermore, we propose a density-matrix-based superclass generation strategy employing eigenvalue decomposition to extract critical quantum state features effectively, enabling a more accurate and robust class representation. Experimental evaluations conducted on standard benchmark datasets demonstrate our approach's superior performance, robustness to noise, and computational feasibility compared to traditional classical and existing quantum-enhanced approaches. The results highlight the practical benefits and potential of Quantum HDC for quantum-enhanced classification tasks and pave the way for future advancements in quantum-inspired computational paradigms.
\end{abstract}

\section{Introduction}
Hyperdimensional Computing~\cite{Kanerva2009Hyperdimensional}, also known as Vector Symbolic Architectures~\cite{gayler2004vector}, is a computational paradigm that leverages high-dimensional  (typically larger than 10,000 dimensions) vectors for information representation, processing and learning. Inspired by cognitive science and neuroscience principles, HDC aims to mimic how the human brain encodes, stores, and retrieves information by operating in high-dimensional space. Instead of using low-dimensional numerical representations, HDC operates on high-dimensional vectors that represent concepts or objects distributively.

Information is processed through algebraic operations. These include bundling (element-wise addition with normalization), binding (component-wise multiplication or XOR), and permutation (reordering of dimensions). These operations allow HDC to efficiently perform classification, pattern recognition, associative memory, and temporal sequence modeling tasks. One of the key advantages of HDC is its robustness to noise, as high-dimensional representations inherently provide redundancy and fault tolerance, making it well suited for real-world applications with noisy data. In addition, 
HDC is computationally efficient because it relies primarily on vector arithmetic rather than complex matrix operations, making it well suited for low-power, real-time, and edge computing scenarios such as embedded systems, IoT devices, and neuromorphic computing platforms. Unlike traditional deep learning methods that require extensive training using backpropagation and gradient descent, HDC can learn through a fast, one-time encoding and updating mechanism, which makes it particularly effective for applications with limited training data.

Although HDC has demonstrated substantial advantages, it still faces serious computing and storage bottlenecks with the increase in data complexity and scale. Therefore, using quantum enhancement to break through the traditional architecture and mechanism of HDC is the motivation for this study.

QeHDC breaks through traditional architectural limitations via three core mechanisms. \textbf{Dimensional Compression:}  Quantum states utilize the geometric properties of the Bloch sphere to represent traditional vector space. By using only $log_2{N}$ qubits, they achieve the same representation capability as classical high-dimension spaces while enabling exponential dimensional compression. \textbf{Computational Architecture:} Quantum Circuit replaces traditional $O(N^2)$ complexity matrix operations with quantum gate operations based on unitary evolution. Through superposition-based parallel processing and entanglement gates, these circuits facilitate efficient feature interactions. \textbf{Representation transformation:} Quantum system overcomes the constraints of classical linear algebra by constructing nonlinear entangled states using quantum circuits. By leveraging quantum interference effects, they achieve superlinear coupling between features, enhancing the interaction dynamics beyond classical approaches. 

Based on the above challenges and motivations, our proposed Quantum-enhanced hyperdimensional computing mainly achieves the following contributions.
\begin{itemize}
    \item First, we introduce a \textbf{two-step hybrid encoder}. It initially maps the original data features into a high-dimensional space using random projection and trigonometric transformations. Then, it expresses the high-dimensional vectors in Bloch space using only a few qubits. This implicitly reduces the dimension of data representation. The proposed encoding method lowers the computational resources overhead required for data processing while making data encoding more suitable for quantum computing implementation.
    \item Second, we develop an innovative \textbf{Quantum Binding} operation called \textbf{Referencec-State-Based Quanutm Binding}. Traditional HDC binding operations often rely on simple $XOR$, circular convolution, or Hadamard product. In contrast, our quantum binding and bundling approaches leverage a predefined quantum reference estate as a binding anchor. By designing specialized quantum circuits, such as the controlled rotation gate (CRX) and the entanglement gate (CX), our method achieves a more efficient data association and representation. This reference-state-based binding not only enhances the expressiveness of high-dimensional vectors but also provides a natural operation mode that is well-suited for quantum hardware implementation.
    \item Third, we propose a Density-Matrix-Based template state construction method, known as \textbf{Density-Matrix-Based Superclass Generation}. This method decomposes the density matrix of quantum states and selects key feature vectors to construct a representative template state (SuperClass) for each class. By capturing core features from the data more accurately, this approach improves the model's robustness and classification accuracy.
    \item Additionally, we built a complete classification experimental verification system based on IBM quantum computing framework Qiskit. First, we complete the classification experimental modeling in high-fidelity quantum noise environment in Qiskit Aer Simulator. We then carry out further confirmatory experiments on a small sample of physical qubits (QPUs) on IBM processors. It is worth noting that despite the essential differences in the calculation principle and noise environment, we still get the corresponding results. This fully proves that this scheme has a unique cross-platform adaptability, which can not only validate and optimize the algorithm in the existing classical computing system but also seamlessly connect with the quantum computing unit in the NISQ era to realize prototype verification.
\end{itemize}
Through these innovations, our QeHDC approach overcomes the efficiency bottleneck of traditional HDC when processing large and complex data. It provides a new idea and a feasible solution for the practical application of quantum computing in machine learning. 

The remaining paper is organized as follows. Section \ref{sec: Backgroud} presents the background, including HDC-related work and quantum computing basics. Section \ref{sec: Quantum-enhanced Hyperdimensional Computing} introduces the overview framework of our proposed method and detailed designs of each component. Section \ref{sec: Experiments} shows the experiments set-up and data in detail. Section \ref{sec: Results and Analysis} provides the results on classification tasks and the comprehensive evaluation of our approach. The final section \ref{sec: Conclusion and Future Work} concludes the paper and envisions the direction of future work.

\section{Backgroud}
\label{sec: Backgroud}
\subsection{Hyperdimensional Computing}
HDC is originally conceived in the 1990s as cognitive models that depend on very high dimensionality and randomness \cite{Kanerva2009Hyperdimensional}. It is inspired by the way the human brain represents and processes information. HDC has the advantages of robustness, fault tolerance, and efficient parallel computing by using high dimensional random vectors to represent and process information \cite{chang2023recent}. In recent years, with the rapid development of artificial intelligence and computing hardware, HDC has regained widespread attention from researchers. In 2017, Rahimi et. al. proposed HDC for machine learning tasks, demonstrating its high efficiency and low complexity in simple text classification \cite{najafabadi2016hyperdimensional}. Subsequently, Hersche et al. evaluated a variety of HDC coding strategies in the motor imagination classification of EEG, demonstrating high energy efficiency on embedded devices\cite{schindler2021primer}.

In recent years, many innovative HDC methods have emerged. AdaptHD \cite{imani2019adapthd} can dynamically adapt to data changes and effectively deal with non-stationary data environments. The real-time incremental learning framework built by OnlineHD \cite{9474107} can support continuous updating of streaming data tasks. In terms of representation optimization, CompHD \cite{8824908} adopts a composite coding strategy to enhance feature interaction, LeHDC \cite{duan2022lehdc} optimizes high-dimensional space mapping through the learnable embedding mechanism, and NeuraldHD \cite{zou2021scalable} deeply integrates a neural network architecture to achieve joint representation optimization. In the field of computational efficiency optimization, the dynamic coding technology designed by DistHD \cite{Wang2023DistHD:} not only improves accuracy but also strengthens noise immunity. QuantHD \cite{imani2019quanthd} adopts low bandwidth representation to significantly reduce the storage overhead on the premise of ensuring performance, while SparseHD \cite{8735551} reduces the computational complexity based on sparse coding theory. For specific application scenarios, BioHD \cite{Zou2022BioHD:} developed a biometric coding protocol to enhance the sensitivity of genome analysis. The graph structure coding operator constructed by GraphHD \cite{nunes2022graphhd} highly improved the accuracy of node classification and SecureHD \cite{10884510} ensured the privacy of medical data processing through the homomorphic encryption protocol. In terms of application, HDC has been widely used in a variety of complex tasks in recent years, including biological signal analysis \cite{eggimann20215, Pale2021Systematic}, natural language processing \cite{Karunaratne2019In-memory}, robot perception and navigation \cite{Menon2022On, Kwon2024Brain-Inspired}, and remote sensing image processing \cite{Mitrokhin2019Learning}.

In summary, HDC is gradually becoming an important research direction in the field of artificial intelligence due to its computational simplicity, robustness, and compatibility with emerging computing architectures, especially in the context of quantum computing convergence, which has a broader prospect.
\subsection{Quantum Computing Basics}
Quantum computing is a computational model based on the principles of quantum mechanics, which uses qubits as the basic computing unit. Unlike classical bits, which exist strictly in either $0$ or $1$, qubits can both be in state $\ket{0}$ and $\ket{1}$. Qubits can exist in a \textbf{superposition}\cite{najafabadi2016hyperdimensional} of both states, represented mathematically as:
\begin{equation}
\ket{\psi} = \alpha\ket{0}+\beta\ket{1},\text{where } \alpha,\beta\in\mathbb{C},|\alpha|^2+|\beta|^2 = 1
\end{equation}

$\alpha$ and $\beta$ are the amplitudes of the complete complex probabilities, indicating the probabilities of obtaining $\ket{0}$ and $\ket{1}$, respectively, when measuring. Due to the existence of superposition states, quantum computing can simultaneously process multiple computation paths in the most exponential level of state space, thus increasing computing power.

Non-local correlations can be formed between multiple qubits through entanglement, where the measurement of one qubit instantaneously affects the state of the other one, even if they are far apart. For example, the \textbf{Bell State}\cite{Allen2015Quantum} of two qubits is one of the most basic entangled states:
\begin{equation}
\ket{\Phi^+}=\frac{\ket{00}+\ket{11}}{\sqrt{2}}
\end{equation}
Entanglement is particularly important in high-dimensional computing because it allows data to form more complex relationships between multi-qubit systems, improving the ability to represent information. 
Quantum computing manipulates qubits through a series of quantum gates, which are unitary transformations that guarantee the reversibility of quantum information. Typical quantum gates include the Hadamard gate (H), Pauli-X (bit-flip) gate, rotation gates (RX, RY, RZ), and controlled gates such as controlled-NOT (CNOT or CX gate), which flip the state of a target qubit conditioned on the state of a control qubit. For example, the widely used H gate generates superposition states from computational basis states:

\begin{equation}
H\ket{0} = \frac{\ket{0}+\ket{1}}{\sqrt{2}},\quad H\ket{1} = \frac{\ket{0}-\ket{1}}{\sqrt{2}}
\end{equation}
Moreover, practical quantum systems typically cannot be described completely by pure states due to noise and environmental interaction. Thus, introduce the concept of \textbf{density matrices}. A density matrix\cite{Thekkadath2016Direct} is a generalized mathematical tool used to represent mixed states (statistical ensembles) of a quantum system:
\begin{equation}
    \rho = \sum_ip_i{\ket{\psi_i}\bra{\psi_i}},\quad \sum_ip_i =  1
\end{equation}
where $p_i$ represents the probability that the quantum system exists in the state $\psi_i$. Density matrices are crucial for quantum computing tasks involving statistical modeling and noisy environments, providing a powerful way to encode complex statistical information beyond pure-state representation. In this paper, the density matrix method is used to construct the Superclass vector, and the core information of the class is extracted by feature decomposition to improve the classification model's generalization ability.
Finally, information encoded in quantum states is obtained through the operation of \textbf{measurement}. Measurement collapses the quantum superposition into one of the classical basis states with probabilities determined by the amplitude squared of each basis state. Specifically, measuring a qubit $\ket{\psi} = \alpha\ket{0}+\beta\ket{1}$ yields outcomes with probabilities:
\begin{equation}
P(\ket{0})=|\alpha|^2, \quad P(\ket{1}) = |\beta|^2   
\end{equation}
In our study, measurement is used in the classification decision process of quantum high-dimensional computation. The data features of hyperdimensional representation are extracted by measurement, and the final pattern recognition task is performed using the classical computational framework.
In summary, quantum computing provides the most new information representation and computation framework; through the superposition of quantum states, entanglement, statistical representation of density matrices, operation of quantum circuits, and measurement mechanisms, it can enhance the feature representation of hyper-dimensional computing. Based on these quantum properties, a novel Quantum-enhanced Hyperdimensional Computing (QeHDC) method is proposed to encode, bind, and classify high-dimensional data more efficiently.

\section{Quantum-enhanced Hyperdimensional Computing}
\label{sec: Quantum-enhanced Hyperdimensional Computing}
\begin{figure}[htbp]

\centering
\resizebox{\textwidth}{!}{
 \includegraphics[scale = 0.6]{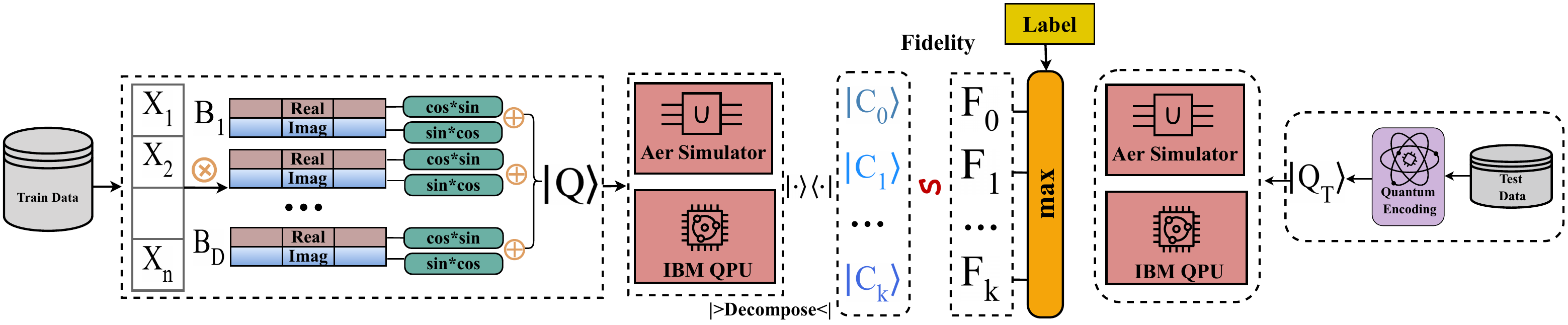}}
\caption{Method}
\label{fig}
\end{figure}
In this paper, we propose a Quantum-enhanced Hyperdimensional Computing (QeHDC) system that leverages quantum computational properties to enhance the efficiency and expressiveness of classical HDC. The system comprises four core modules. First, a 2-step classical-quantum hybrid encoder transforms classical data into quantum states. Then, a Quantum HD Operator performs HDC operations like binding and construction using quantum circuits. Superclass construction builds a density matrix over bound states and extracts features via eigen-decomposition. Finally, classification is achieved by comparing input states with learned superclasses using quantum state fidelity. We will introduce each of these parts in detail below.
Before that, we define a system input sample as:
\begin{equation}
D = {(X_i,y_i)}_{i=1}^N
\end{equation}
where $X_i\in\mathbb{R}^d$ is the feature vector of dimension $d$ and $y_i$ is the class label, typically encoded as a categorical variable.
\subsection{2-step Hybrid Encoder}
The core of the coding process is to efficiently map classical eigenvectors to quantum states while preserving the structural information of the input data. The first step is a classical transformation. Given a classical input vector $X\in\mathbb{R}^d$, we perform a random projection using a randomly initialized weight matrix $W\in\mathbb{R}^{n*d}$, where $n = 2^{N_{\textbf{qubits}}}$ represents the target quantum dimension. The transformation is computed as:
\begin{equation}
Z = WX
\end{equation}
To introduce nonlinearity and phase variation in the coding process, we introduce a bias term $B\in\mathbb{R}^n$ whose values are uniformly sampled from the interval $[0,2\pi]$ forming a transformed projection:
\begin{equation}
    Z' = Z+B
\end{equation}
To convert the projected values into \textbf{quantum-compatible} complex-valued presentations, we 
Then, the real part and imaginary part are mapped separately using the sinusoidal transformations:
\begin{equation}
    \Phi_{\text{re}} = cos(Z')sin(Z),\quad \Phi_{\text{im}} = sin(Z')cos(Z)
\end{equation}
This formulation which we called \textbf{Cross-Multiplicative} 
achieves the same uniform phase angle distribution while using only single weight matrix, making it the most efficient approach (See Appendix A for explanation) The final quantum state representation is given by:
\begin{equation}
\Psi = \frac{\Phi_{\text{re}}+i\Phi_{\text{im}}}{||\Phi_{\text{re}}+i\Phi_{\text{im}}||}
\end{equation}
Compared to other encoding schemes, this approach ensures a well-distributed representation in the complex space, enhancing separability and stability in quantum computation. Further theoretical analysis, including its impact on data distribution, is provided in Appendix \ref{appendix:parameterusage}.
\subsection{QeHDC Operation}
Quantum operations form the core of our method by enabling transformations and interactions between encoded quantum states. Once classical data is converted into quantum representations via Cross-Multiplicative Encoding, a set of quantum operations is applied to enhance, bind, and extract relevant information. The goal of these operations is to exploit quantum properties such as superposition and entanglement to improve computational efficiency and feature representation.
We focus on the \textbf{binding operation}, which is critical in representing relationships between encoded quantum states. Unlike classical models that rely on matrix multiplications or convolutional operations, our approach leverages quantum state interactions to achieve a compact and efficient representation of information.

In the Quantum hyperdimensional system, direct implementation of classical HDC binding operation causes an exponential increase in the required qubits (See Appendix \ref{appendix:compressedbinding}). To address this, we propose a specifically compressed binding method, employing a quantum circuit for efficient information binding and compression.
Consider the encoded quantum state for a single input sample as :
\begin{equation}
\ket{\Phi_{\text{in}}} = \sum_{i=0}^{2^N-1}\alpha_i\ket{i},\quad \sum_i|\alpha_i|^2 =1
\end{equation}
Also define a predefined reference quantum state as:
\begin{equation}
   \ket{ \Phi_{\text{ref}}} = \sum_{j=0}^{2^N-1}\beta_j\ket{j},\quad \sum_j|\beta_j|^2 =1
\end{equation}

We initialize these two states into two separate quantum registers, each containing $n$ qubits, resulting in the initial state:
\begin{equation}
    \ket{\Psi_{\text{init}}} = \ket{\Phi_{\text{in}}}\otimes\ket{\Phi_{
    \text{ref}}}
\end{equation}
Next, we apply Controlled-RX (CRX) gates between each pair of corresponding qubits ($q_i, q_{i+n}$) with rotation angle $\pi/4$:
\begin{equation}
    CRX(\pi/4) = \ket{0}\bra{0}\otimes{I}+\ket{1}\bra{1}\otimes{e^{-i\frac{\pi}{8}X}}
\end{equation}
At this point, the state evolves into:
\begin{equation}
    \ket{\Psi'} = \prod_{i=0}^{n-1}CRX(\pi/4)\ket{\Psi_{\text{init}}}
\end{equation}
Subsequently, apply single-qubit RY rotations on each qubit in the reference register for additional nonlinear compression:
\begin{equation}
    RY(\pi/4) = e^{-i\frac{\pi}{8}Y}
\end{equation}
The state then becomes:
\begin{equation}
    \ket{\Psi''} = \prod_{i=0}^{n-1}RY_i(\pi/4)\ket{\Psi'}
\end{equation}
To further enhance the binding and compression effect, we add chain entanglement operations (CX gates) within the reference register. The final highly entangled and compressed quantum representation is:
\begin{equation}
    \ket{\Psi_{\text{comp}}} = \prod_{\{q_a,q_b\}\in{Q}}CX_{q_a,q_b}\ket{\Psi''}
\end{equation}
\begin{equation}
    \ket{\Psi_{\text{comp}}} = U\ket{\Psi_\text{init}}
\end{equation}
Where $ Q=\{n,n+1,...,2n-1\}$.
For all samples within the same class $c$ (assuming a total of $m$ samples), we perform the above binding and compression process individually, obtaining:
\begin{equation}
    \{\ket{\Psi_{\text{comp}}^{(1)}},\ket{\Psi_{\text{comp}}^{(2)}},...,\ket{\Psi_{\text{comp}}^{(m)}}\}
\end{equation}
The binding circuit described above is a fully entangled structure, meaning any two qubits in the latter half are directly entangled. The circuit for the example with 4 qubits is shown in Fig. 3.

\begin{figure}[ht]
\centering
\resizebox{\textwidth}{!}{\begin{quantikz}[row sep=0.3cm, column sep=0.4cm]
\lstick{$q_0$} & \ctrl{4} & \qw & \qw & \qw & \qw& \qw& \qw& \qw& \qw& \qw& \qw& \qw\\
\lstick{$q_1$} &\qw& \ctrl{4} & \qw & \qw & \qw& \qw& \qw& \qw & \qw& \qw& \qw& \qw\\
\lstick{$q_2$} &\qw&\qw& \ctrl{4} & \qw & \qw & \qw & \qw& \qw& \qw& \qw& \qw& \qw\\
\lstick{$q_3$} &\qw&\qw&\qw& \ctrl{4} & \qw & \qw & \qw & \qw& \qw& \qw& \qw& \qw\\
\lstick{$q_4$} & \gate{R_x(\pi/4)} &\qw&\qw&\qw& \gate{R_y(\pi/4)} & \ctrl{1}&\ctrl{2}&\ctrl{3} & \qw& \qw& \qw& \qw \\
\lstick{$q_5$} &\qw& \gate{R_x(\pi/4)} &\qw&\qw& \gate{R_y(\pi/4)} & \targ{}& \qw& \qw & \ctrl{1}&\ctrl{2}& \qw& \\
\lstick{$q_6$} &\qw&\qw& \gate{R_x(\pi/4)} &\qw& \gate{R_y(\pi/4)} & \qw&\targ{}& \qw&\targ{}& \qw& \ctrl{1} & \qw\\
\lstick{$q_7$} &\qw&\qw&\qw& \gate{R_x(\pi/4)} & \gate{R_y(\pi/4)} & \qw& \qw& \targ{} & \qw &\targ& \qw& \targ{}& \qw
\end{quantikz}}
\caption{Quantum Bind Circuit}
\label{fig:quantum-circuit}
\end{figure}
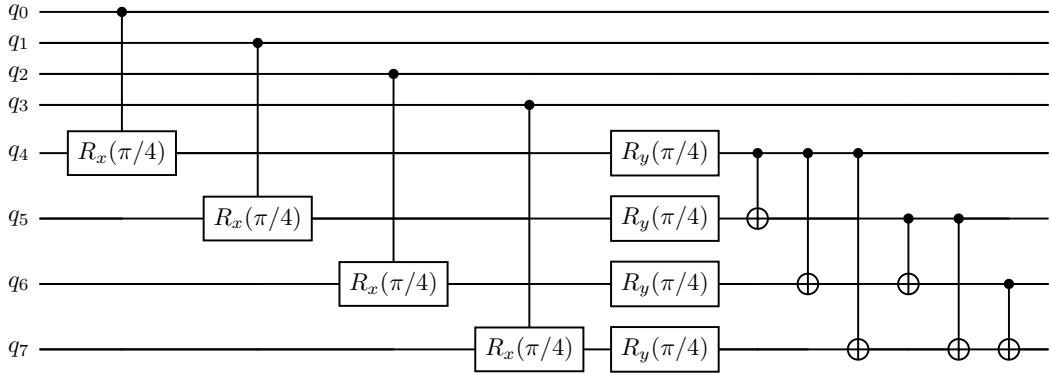

We also employed some structures with lower complexity. Please refer to Appendix \ref{appendix:bindcircuits}.
Through these detailed steps, we achieve quantum state binding and compression at both sample and class level, significantly avoiding the exponential qubit resource increase caused by the traditional classical HDC method. This approach holds substantial potential and expansion possibilities in future practical applications.
In our proposed compressed binding method, we implement quantum state binding compression using specifically designed quantum circuits. In our practical study, we employ two approaches for realizing and verifying our quantum bound states:
\begin{itemize}
    \item Ideal Quantum State simulation via the Qiskit Aer Simulator (no measurement required)
    \item Physical experiments on IBM's Quantum Processing Unit (QPU), requiring measurement and quantum state tomography for reconstruction. 
\end{itemize}
It should be noted that using the ideal state vector from the Aer simulator applies only to theoretical analysis and algorithm validation. On real quantum devices (e.g., IBM QPU), measurement and quantum state tomography reconstruction are essential steps. Our research encompasses both ideal simulations and real-device experiments to comprehensively demonstrate both the theoretical performance and the practical feasibility of our quantum binding algorithm.
\subsection{SuperClass Template Generate}
In our method, using the previously introduced compressed binding method, each class contains numerous samples represented by bound quantum states. Therefore, from multiple sample states with the same class, we must extract a representative and stable quantum state, termed the \textbf{SuperClass Template State}.  
Suppose a class $c$ has $m$ data samples, after obtaining a holistic representation for the class, we need to comprehensively capture the statistical characteristics of these sample states. The quantum density matrix provides the ideal mathematical tool for this purpose:
\begin{equation}
    \rho_c = \frac{1}{m}\sum_{i=1}^{m}\ket{\Psi_{\text{comp}}^{i}}\bra{\Psi_{\text{comp}}^{i}}
\end{equation}
The density matrix $\rho_c$ is essentially a statistical representation of all bound states in class $c$. It reflects information from all states and reveals intrinsic correlations among them. The density matrix has dimensions $2^{2n}\times2^{2n}$. To address the computational challenge due to dimensionality, we employ two strategies based on matrix dimension. Directly construct the density matrix using standard methods and perform eigen-decomposition:

\begin{equation}
     \rho_c\ket{\phi_j}=\lambda\ket{\phi_j}
\end{equation}
We choose the eigenvector corresponding to the largest eigenvalue $\lambda_{\text{max}}$ as the representative state for the class:
\begin{equation}
    \ket{\Phi_c^{\text{template}}} =  \ket{\phi_{\text{max}}},\quad\rho_c\ket{\phi_{\text{max}}}=\lambda\ket{\phi_{\text{max}}}
\end{equation}
Through the described SuperClass template state generation approach, we effectively extract representative and stable class-level quantum representations from multiple bound states.
\subsection{Single-Pass Training}
In our method, we use one-pass training. Thanks to the characteristics of the high-dimensional computing framework, the model can be constructed with only one data traversal, avoiding the computational cost and parameter tuning complexity caused by multiple iterative optimizations in traditional machine learning methods. This approach not only significantly reduces the occupation of quantum resources, but is more suitable for the current environment with limited quantum hardware resources, and effectively improves the actual deployment capability and generalization performance of the algorithm.
\subsection{Classification}
In the classification phase, for each unseen classical data sample $X_{\text{test}} $, we first encode it into a quantum state $\ket{\Phi_{X_{\text{test}}}} $ using the Cross-Multiplicative Encoding approach described previously. Subsequently, the encoded state is bound to the same predefined reference quantum state $|\Phi_{\text{ref}}\rangle $ via the compressed Binding process, obtaining a quantum bound state:
$$
|\Psi_{\text{test}}\rangle = \mathcal{U}_{\text{bind}}\left(|\Phi_{x_{\text{test}}}\rangle, |\Phi_{\text{ref}}\rangle\right).
$$
Classification is then performed by calculating the quantum state fidelity between the bound test state $\ket{\Psi_{\text{test}}}$ and each pre-computed SuperClass template state $ \ket{\Psi_{c_j}^{(\text{template})}} $:
$$
F\left(|\Psi_{\text{test}}\rangle, |\Psi_{c_j}^{(\text{template})}\rangle\right) = \left|\langle\Psi_{\text{test}}|\Psi_{c_j}^{(\text{template})}\rangle\right|^2.
$$
The class prediction for the test sample is obtained by selecting the SuperClass template that yields the highest fidelity value:
$$
c^* = \arg\max_{c_j} F\left(|\Psi_{\text{test}}\rangle, |\Psi_{c_j}^{(\text{template})}\rangle\right).
$$
On ideal quantum simulators (e.g., Qiskit Aer), the fidelity is directly computed using state vectors, whereas on real quantum hardware (e.g., IBM QPUs), quantum state tomography is first performed to reconstruct the density matrix $\rho_{\text{test}} $. Subsequently, fidelity is calculated as:
$$
F(\rho_{\text{test}}, |\Psi_{C_j}^{(\text{template})}\rangle) = \langle\Psi_{C_j}^{(\text{template})}|\rho_{\text{test}}|\Psi_{C_j}^{(\text{template})}\rangle.
$$

This quantum fidelity-based classification approach naturally leverages quantum coherence and quantum state similarity, providing robust, noise-resistant classification performance that is particularly suitable for resource-constrained quantum computing environments.

\section{Experiments}
\label{sec: Experiments}
In this section, we evaluate the performance and effectiveness of our proposed QeHDC framework. We provide details on the datasets, quantum simulation, and hardware environments, as well as the parameter settings shared across all experiments.
\subsection{Dataset}
We evaluate our method on three widely used datasets: \textbf {ISOLET}\cite{isolet_54}, \textbf{MNIST}\cite{726791}, and \textbf{UCI HAR}\cite{8567275}. The quantitative details of the datasets are shown in Table \ref{tab:data detail}.
\begin{table}[htbp]
 \setlength{\tabcolsep}{2pt}
\caption{Quantitative details of the datasets}
\begin{center}
\begin{tabular}{c|c|c|c|c}
\toprule
 Dataset& \# Features & \# Classes &\# Training Samples &\# Test Samples\\
\midrule
MNIST&784&10&60000&10000\\
\midrule
ISOLET&617&26&6238&1559\\
\midrule
UCI HAR&561&6&7352&2947\\
\bottomrule
\end{tabular}
\label{tab:data detail}
\end{center}
\end{table}

The ISOLET dataset consists of spoken letter recordings, where each sample corresponds to one of the 26 English alphabet letters spoken by different speakers. Each instance is represented by 617 acoustic features. Due to its high-dimensional input and large class space, ISOLET presents a challenging scenario for scalable classification.
\begin{table}[htbp]
\centering
\footnotesize
\caption{Binary Classification Results (2 Classes) for Three Datasets by Method and Dimension}
\label{tab:binary_all}
\resizebox{\textwidth}{!}{
\begin{tabular}{l|ccc|ccc|ccc}
\toprule
\multirow{2}{*}{Method} & \multicolumn{3}{c|}{ISOLET-2Classes} & \multicolumn{3}{c|}{MNIST-2Classes} & \multicolumn{3}{c}{UCI HAR-2Classes} \\
\cmidrule(lr){2-4} \cmidrule(lr){5-7} \cmidrule(lr){8-10}
                        & 16D & 32D & 64D & 16D & 32D & 64D & 16D & 32D & 64D \\
\midrule
Vanilla   & 72.78$\pm$8.12  & 76.11$\pm$5.11  & 76.39$\pm$4.37  & 78.50$\pm$8.90  & 90.12$\pm$2.04  & 94.47$\pm$0.92  & 60.15$\pm$1.64  & 67.63$\pm$2.12  & 70.70$\pm$2.79 \\
AdaptHD   & 67.50$\pm$13.96 & 68.33$\pm$6.49  & 79.72$\pm$8.20  & 77.37$\pm$8.20  & 90.10$\pm$2.82  & 96.71$\pm$0.83  & 57.15$\pm$1.96  & 62.50$\pm$8.14  & 70.29$\pm$1.45 \\
OnlineHD  & \textbf{88.89$\pm$4.53}  & 95.28$\pm$0.39  & \underline{98.89$\pm$1.04}  & \textbf{95.19$\pm$1.76}  & \textbf{99.16$\pm$0.34}  & \underline{99.67$\pm$0.12}  & \textbf{87.80$\pm$4.17}  & \underline{88.38$\pm$4.15}  & 90.90$\pm$1.03 \\
NeuralHD  & \underline{86.39$\pm$2.58}  & \underline{95.83$\pm$1.80}  & 96.39$\pm$0.79  & \underline{94.44$\pm$1.03}  &\underline{98.49$\pm$0.38}  & 99.01$\pm$0.32  & 83.52$\pm$2.98  & 87.56$\pm$2.56  & \underline{91.97$\pm$1.35} \\
CompHD    & 55.83$\pm$5.14  & 68.89$\pm$1.57  & 75.83$\pm$4.76  & 72.37$\pm$12.39 & 87.42$\pm$7.31  & 90.76$\pm$2.61  & 57.05$\pm$5.21  & 68.05$\pm$2.05  & 64.84$\pm$4.27 \\
SparseHD  & 66.94$\pm$1.42  & 61.94$\pm$4.43  & 70.83$\pm$4.25  & 68.31$\pm$17.31 & 72.58$\pm$2.37  & 66.49$\pm$9.15  & 57.33$\pm$2.83  & 62.84$\pm$6.95  & 68.39$\pm$4.71 \\
QuantHD   & 55.00$\pm$3.60  & 75.56$\pm$6.85  & 74.17$\pm$1.36  & 70.50$\pm$19.59 & 83.04$\pm$0.60  & 82.81$\pm$6.94  & 55.33$\pm$5.70  & 61.67$\pm$12.14 & 67.56$\pm$6.21 \\
LeHDC     & 59.44$\pm$5.15  & 56.39$\pm$1.04  & 65.00$\pm$5.31  & 51.06$\pm$5.21  & 61.61$\pm$1.58  & 67.30$\pm$11.97 & 52.36$\pm$2.04  & 55.05$\pm$0.83  & 53.43$\pm$2.39 \\
\midrule
&4 Qubits&5 Qubits&6 Qubits&4 Qubits&5 Qubits&6 Qubits&4 Qubits&5 Qubits&6 Qubits\\
\midrule
Ours-Aer & 86.35$\pm$1.11 & \textbf{96.05$\pm$0.57} & \textbf{98.97$\pm$0.49} & 89.99$\pm$0.62 & 95.61$\pm$1.34 & \textbf{99.97$\pm$0.83} & \underline{87.59$\pm$1.36} & \textbf{89.72$\pm$1.43} & \textbf{92.04$\pm$0.26} \\
\bottomrule
\end{tabular}
}
\end{table}
The MNIST dataset is a benchmark image classification task composed of handwritten digits from 0 to 9. Each image is 28×28 pixels, flattened into a 784-dimensional vector.  MNIST is widely used for evaluating pattern recognition and machine learning algorithms.

The UCI HAR (Human Activity Recognition) dataset contains 561-dimensional feature vectors extracted from smartphone sensor signals. The data represents six human activities (e.g., walking, standing, lying down) collected from multiple subjects. It is a compact but structured dataset that tests the model's ability to distinguish between temporal movement patterns.
These datasets vary in feature dimension, number of categories, and sample complexity, providing different assessment environments. To investigate the extensibility of our model, we performed a classification task ranging from 2 classes to 10 classes, rather than using a complete set of classes (for example, all 26 letters in ISOLET or all 10 numbers in MNIST). We observed that the full class classification task led to a significant decline in performance, so we adopted a progressive evaluation strategy that better reflects the actual capabilities of current quantum systems.
\subsection{Low-Dimensional HDC and Quantum Encoding}
While our approach is inspired by traditional HDC, we intentionally work in a significantly lower dimension([16,32,64]) to reflect the limitations of current quantum hardware. In the classical HDC literature, extremely high sizes (up to 10,000) are often used to ensure robustness and orthogonality. In contrast, our QeHDC performs well in low dimensions using the expressiveness of quantum states. For fair comparison, all HDC baselines are also constrained to the same vector dimension. This highlights a key difference in the design. While traditional techniques rely on the force dimension, our approach seeks to preserve a quantum-efficient representation of core HDC principles.
\subsection{Implementation Details}
Experiments are conducted in the following environments:
\begin{itemize}
    \item \textbf {Qiskit Aer Simulator:}\cite{qiskit2024}  An ideal, noiseless quantum simulation environment used for baseline accuracy and fidelity evaluation. The majority of our experiments are conducted using the Aer simulator, as it allows scalable, noise-free validation of our method across different datasets and class counts.
    \item \textbf{IBM QPU:}~\cite{Sturm2023Theory}A small scale of experiments is deployed on real IBM QPU backends to assess the practical feasibility of our framework under hardware constraints. Quantum state tomography is used to reconstruct output states due to the non-availability of state vectors on real hardware. Basic error mitigation techniques are applied where applicable. For comparison, the same scale and data samples used on IBM devices are also evaluated using the Aer simulator to establish a controlled benchmark.
    \item \textbf{Classical Baselines}: For traditional HDC and classical parts of our methods, we use TorchHD \cite{heddes2023torchhd} implementations. All classical experiments are run on a workstation equipped with RTX A6000, GPU acceleration is used for efficient batching in TorchHD. Dimensionality (16, 32, 64) is matched with the quantum model, and all baselines are evaluated on the same training/testing splits as the quantum system.
    \end{itemize}
All quantum operations, including encoding, binding, and fidelity computation, are implemented using Qiskit. State preparation and circuit compilation are optimized for each backend. In real-device executions, we use 8192 shots and transpile circuits to the native gate set of the selected backend. Fidelity measurements are obtained by reconstructing the output density matrices via tomography.
Hyperparameters, such as the number of qubits (\{4,5,6\}), repeats (5), and choice of reference state, are fixed across experiments unless specified in dedicated analysis sections.
\section{Results and Analysis}
\label{sec: Results and Analysis}
In this section, we present the key findings from our experiments. We focus on evaluating classification accuracy and the contribution of each core component in the QeHDC framework. All results are organized by task complexity and dataset, followed by comparative studies to highlight the advantages and limitations of our approach.
\subsection{Overall Performance}
Table \ref{tab:binary_all}  provides a comprehensive comparison of our QeHDC framework against various classical hyperdimensional computing (HDC) baselines—such as VanillaHD, AdaptHD, OnlineHD, NeuralHD, CompHD, SparseHD, QuantHD, and LeHDC—across three benchmark datasets (ISOLET, MNIST, and UCI HAR) under different dimensional constraints (16D, 32D, and 64D). The dimensions in the classical methods correspond respectively to quantum states encoded by 4, 5, and 6 qubits in our approach.

From these results, our QeHDC consistently achieves robust and superior classification accuracy. Specifically, on the ISOLET binary classification task, our model demonstrates accuracy that increases significantly with the dimension, reaching 98.97\% at 6 qubits (64D), clearly outperforming all classical baselines except OnlineHD, with which it achieves comparable performance. In the MNIST binary classification scenario, our approach further excels, reaching an accuracy of 99.97\% at 6 qubits—surpassing or matching the best classical results. Likewise, for the UCI HAR dataset, which presents complex temporal feature patterns, our QeHDC method attains an accuracy of 92.04\% at 6 qubits, which notably exceeds most classical baselines.

Importantly, even at lower-dimensional settings , our method maintains competitive performance. This highlights the effective expressive capacity of our quantum-inspired encoding and compression-based binding mechanisms. Such robust performance under dimensional constraints is particularly relevant for near-term quantum computing applications, where resource availability (e.g., number of qubits) is significantly limited.

Beyond binary classification tasks, our method also demonstrates promising results in more complex multi-class classification settings, systematically evaluated from 3-class up to 10-class problems. These additional experiments confirm that the QeHDC framework consistently retains strong classification capability across increasing classification complexity and various datasets. Detailed multi-class classification results and further analysis can be found in the Appendix \ref{appendix:results}. 

Overall, these extensive experimental evaluations indicate that our quantum-based HDC approach not only competes favorably against classical HDC methods but also shows distinct advantages regarding efficiency, scalability, and robustness across different classification scenarios.
\subsection{Limitation Analysis}

\begin{figure}[ht]
\centering
\begin{subfigure}{0.45\linewidth}
    \includegraphics[width=\linewidth]{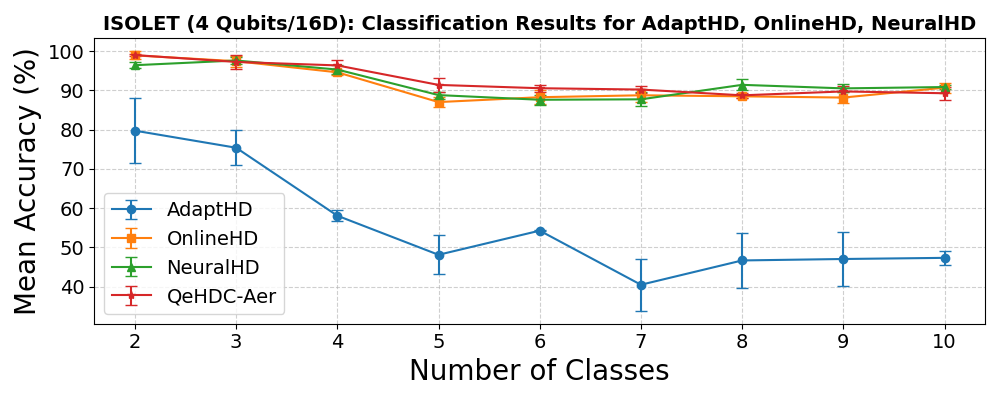}
    \caption{Performance on ISOLET Dataset}
    \label{fig: isolet analysis}
\end{subfigure}
\hspace{0.05\linewidth}
\begin{subfigure}{0.45\linewidth}
    \includegraphics[width=\linewidth]{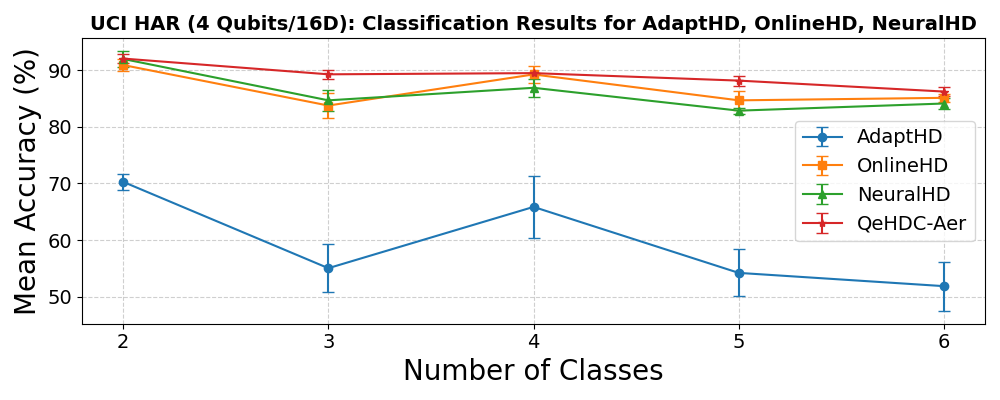}
    \caption{Performance on UCI HAR}
    \label{fig: ucihar analysis}
\end{subfigure}
\caption{Mean classification accuracy of AdaptHD, OnlineHD, NeuralHD, and QeHDC-Aer on (a) ISOLET and (b) UCI HAR datasets with varying numbers of classes.}
\label{fig:two_images_subcaption}
\end{figure}

From the figures Fig. \ref{fig: isolet analysis} and Fig. \ref{fig: ucihar analysis}, it is evident that our proposed Quantum Hyperdimensional Computing framework (QeHDC-Aer) consistently demonstrates stable and superior classification accuracy, around 90\%, across varying classification complexities (from 2-class up to 10-class scenarios) on both ISOLET and UCI HAR datasets. In contrast, classical methods such as AdaptHD experience significant accuracy degradation and instability as the number of classes increases, highlighting the robustness of our approach under more complex classification tasks. Moreover, our method achieves performance comparable to, or better than, top-performing classical methods like OnlineHD and NeuralHD, especially notable on the UCI HAR dataset. Other classical baselines (e.g., VanillaHD, CompHD, SparseHD) fail to maintain satisfactory performance at such low-dimensional settings, indicating difficulty in handling complex tasks under limited dimensional constraints. These results clearly demonstrate the strong generalization capability and scalability of our QeHDC framework, emphasizing its advantage over classical HDC approaches in low-dimensional, resource-constrained quantum environments.
\subsection{Aer Simulation and IBM QPU}
\begin{table}[htbp]
 \setlength{\tabcolsep}{2pt}
\caption{Small-batch Classification on UCI HAR (4 
 Qubits)}

\begin{center}
\begin{tabular}{c|c|c|c|c|c}
\toprule
 Method& \# Qubits &\# Train & \# Test   &Accuracy & Fidelity (to Aer Ideal)\\
\midrule
QeHDC-Aer&4&50&20&0.80&0.98\\
\midrule
QeHDC-QPU&4&50&20&0.75&0.87\\

\bottomrule
\end{tabular}
\label{tab:Aer and QPU}
\end{center}
\end{table}
To evaluate both the ideal performance and practical feasibility of our proposed QeHDC framework, we conduct experiments using the Qiskit Aer simulator and IBM Quantum Processing Units (QPUs). The Aer simulator provides a noise-free quantum environment with access to full quantum statevectors, allowing us to assess the expressiveness and fidelity of our model under ideal conditions, free from decoherence and gate imperfections. In contrast, IBM QPUs introduce realistic limitations such as noise, qubit connectivity constraints, and readout errors, offering a platform to validate the framework under physical quantum computing constraints.

On the Aer simulator, we observed high fidelity and classification accuracy for the quantum states generated after binding and template construction. These results validate the theoretical soundness of our encoding, binding, and template generation procedures. Specifically, the fidelity between the ideal post-binding statevector and the density matrix reconstructed via simulated tomography consistently exceeded 0.98, confirming the accuracy and stability of our tomography pipeline in a noise-free setting.

Quantum State Tomography (QST) plays a critical role in real-device testing where full access to quantum states is unavailable. By preparing and measuring the same quantum state across multiple bases and statistically reconstructing its density matrix, QST enables us to recover the output state from QPU measurements and perform fidelity-based classification. In our workflow, the test state is first encoded and the binding circuit is executed on the QPU. The resulting measurement data is then used to reconstruct a density matrix via QST, from which we extract a quantum representation that is compared with the stored SuperClass template to perform final classification.

Due to the significant cost and scaling complexity of QST and QPU, where the number of measurement circuits increases exponentially with the number of qubits. We adopted a small-batch testing strategy. Combined with device runtime constraints, this makes full-scale experiments infeasible on current hardware. As a representative case, we selected a small batch from the UCI HAR dataset (4-qubit setting), training on 50 samples and testing on 20. We redo this experiment for 5 times and get the average results in Table \ref{tab:Aer and QPU}.

These results indicate that even under real noise, the quantum states reconstructed from the QPU retain sufficient structure to support meaningful classification, with fidelity and accuracy close to ideal conditions. Notably, the difference in fidelity between the QPU and simulator also quantifies the impact of hardware noise on the final inference performance.

The examples of our binding circuits, representative tomography circuits, and visual comparisons of ideal and reconstructed states are included in the Appendix \ref{appendix:tomography}.

Overall, this experiment validates the expressiveness of our method under ideal conditions and its deployability on current meso-scale noisy quantum devices (NISQs), providing a practical basis and experimental experience for further expanding the QeHDC model on more powerful quantum hardware platforms in the future.
 \subsection{Ablation Study}
In order to evaluate the effect of coding mechanism on the overall performance of QeHDC systems, we conduct an ablationon on Quantum encoder. The objective of this study is to analyze the effects of different coding strategies on the quality of class templates and the final classification accuracy.
\section{Conclusion and Future Work}
\label{sec: Conclusion and Future Work}
In this paper, a new framework Quantum-enhanced hyperdimensional computing (QeHDC) is proposed, which combines the basic idea of traditional hyperdimensional computing (HDC) with the ability to express quantum states. We design a classic-quantum two-step hybrid coding mechanism and introduce a reference state-based compression binding method to efficiently generate a compact high-dimensional quantum representation that can be used for classification tasks. At the same time, we construct a SuperClass template by means of density matrix decomposition, so that class abstraction can be realized in quantum representation space.

Unlike traditional HDC models, which typically rely on tens of thousands of dimensional vectors, our quantum scheme is able to operate in significantly low-dimensional (e.g. 64-dimensional) spaces, and achieves competitive and even superior results on multiple classification tasks. Experiments on ISOLET, MNIST, and UCI HAR datasets demonstrate the effectiveness and robustness of the proposed method in simulated environments (Aer) and real devices (IBM QPU). Especially in the small batch test of QPU, the model can still obtain high quantum state fidelity and classification accuracy even if there is hardware noise, which indicates that the method has good usability and deployment potential.

In the future, we will promote our work from the following directions: On the one hand, we will expand this method to more complex task scenarios (such as sequence modeling, online learning), and explore the combination of quantum binding and class incremental update mechanisms. On the other hand, to further optimize the computational cost of quantum state chromatography, or to design a more suitable reasoning mechanism for the hardware structure to adapt to larger-scale quantum systems. In addition, the introduction of learnable quantum encoders and differentiable template generation processes is also expected to further enhance the integration of QeHDC with deep architectures such as neural networks. With the continuous development of quantum hardware, we believe that the framework proposed in this work will show its broad application prospects in larger-scale scenarios and provide a solid foundation for building the next generation of quantum cognitive computing systems.
\bibliographystyle{plain}
\bibliography{references}
\clearpage

\appendix
\section{Parameter Usage Comparison}
\label{appendix:parameterusage}
In QeHDC, the choice of encoding strategy significantly affects the \textbf{phase angle distribution} in the complex plane. It directly impacts the separability and expressiveness of quantum states. This section compares and analyzes three sinusoidal transformation strategies: Single-Weight Cross-Multiplicative (our method), Dual-Weight Same-Product, and Single-Weight Same-Product.
We compare these methods with respect to parameter efficiency, phase angle distribution, and normalization properties. 
\subsection{Phase Angle Concentration in Single-Weight Same-Product}
In \textbf{Single-Weight Same-Product Encoding}, both the real and imaginary components share the \textbf{same weight matrix}, defined as:
\begin{equation*}
\text{Re}(\Psi') = \cos(WX + B) \cdot \sin(WX),
\end{equation*}
\begin{equation*}
\text{Im}(\Psi') = \cos(WX + B) \cdot \sin(WX)
\end{equation*}
Since the transformation for both components is identical, their ratio remains constant:
\begin{equation*}
\frac{\text{Im}(\Psi')}{\text{Re}(\Psi')} = 1
\end{equation*}
This results in all encoded data points having the same phase angle. Consequently, the data distribution in the complex plane becomes \textbf{highly concentrated}, reducing the effectiveness of encoding.
\subsection{Parameter Efficiency in Cross-Multiplicative Transformation}
Single-Weight Cross-Multiplicative Encoding solves this issue by using a single weight matrix while applying different transformations to the real and imaginary components:
\begin{equation*}
\text{Re}(\Phi) = \cos(WX + B) \cdot \sin(WX),
\end{equation*}
\begin{equation*}
\text{Im}(\Phi) = \sin(WX + B) \cdot \cos(WX)
\end{equation*}
This ensures a uniform phase angle distribution while minimizing parameter usage.

In contrast, Dual-Weight Single-Product Encoding requires two independent weight matrices to achieve the same effect:
\begin{equation*}
\text{Re}(\Psi'') = \cos(W_{\text{re}} X + B) \cdot \sin(W_{\text{re}} X),
\end{equation*}
\begin{equation*}
\text{Im}(\Psi'') = \cos(W_{\text{im}} X + B) \cdot \sin(W_{\text{im}} X)
\end{equation*}
While this method also maintains a uniform phase angle distribution, it doubles the parameter count, increasing computational overhead.

\section{Compressed Binding}
\label{appendix:compressedbinding}
This appendix provides a detailed analysis demonstrating that traditional binding methods used in classical HDC, such as element-wise multiplication or circular convolution, lead to an exponential explosion in the number of required qubits when directly applied to quantum systems. Although theoretically feasible to use quantum state amplitude as classical high-dimensional vectors for binding, such an approach is practically infeasible in quantum hardware. Additionally, this approach has been explored in previous studies, but predominantly through classical binding in classical HDC, which is usually defined via element-wise multiplication or XOR operation simulations rather than quantum implementation.
Consider 2 $D$-dimensional vectors:  
\begin{equation*}
    A =[a_1,a_2,..,a_D],B =[b_1,b_2,...,b_D]
\end{equation*}  
Their binding result (element-wise multiplication) is expressed as:  
\begin{equation*}
    C=A\circ B = [a_1b_1,a_2b_2,...,a_Db_D]
\end{equation*}Consider 2 $D$-dimensional  vectors:
\begin{equation*}
    A =[a_1,1_2,..,a_D],B =[b_1,b_2,...,b_D]
\end{equation*}
Their binding result (element-wise multiplication) is expressed as:
\begin{equation*}
    C=A\circ B = [a_1b_1,a_2b_2,...,a_Db_D]
\end{equation*}
This type of binding operation is efficient and easy to perform in classical system, but if it is implemented directly in a quantum system, it faces huge resource challenges.

In quantum system, a quantum state of $n$ qubits represents a $2^n$-dimensional complex vector:
\begin{equation*}
    \ket{\psi}=\sum_{i=0}^{2^n-1}\alpha \ket{i},\quad\sum{|\alpha|^2}=1
\end{equation*}
If we want to bind two quantum states using the traditional HDC binding method, we first need to simultaneously represent two full $D=2^n$-dimensional quantum states in the quantum computer. Each stare requires $n$ qubits and the joint representation of two states thus requires 2$n$ qubits. If we bind $k$ quantum states, the toatal number of qubits required scales as:$$Q_{total}=k\cdot n$$ This implies that if we directly adopt traditional binding methods, adding each additional state linearly increases the required qubits. Given that each quantum state's dimensionality grows exponentially, total computational resources effectively explode exponentially. Theoretically, we could directly use quantum state amplitudes as classical high-dimensional vectors for binding.
However, this approach does not genuinely leverage quantum computing advantages, as it simply treats quantum amplitudes as classical data. Hence, it deviates from our quantum algorithmic context, failing to harness quantum parallelism or the computational benefits of quantum superposition.
It is worth noting that previous studies have explored this "amplitude-as-vector" binding approach. Nevertheless, these studies typically relied on classical simulations rather than actual implementations on real quantum hardware.
\section{Bind Circuits Alternatives}
\label{appendix:bindcircuits}
Below, we describe two alternative entanglement structures of bind Circuits with lower complexity than the fully-connected entanglement structure.
\subsection{Linear (Chain) Entanglement}
The circuit for the example with four qubits is shown in the figure.

\begin{figure}[ht]
\centering
\resizebox{\textwidth}{!}{\begin{quantikz}[row sep=0.3cm, column sep=0.4cm]
\lstick{$q_0$} & \ctrl{4} & \qw & \qw & \qw & \qw& \qw& \qw& \qw& \qw\\
\lstick{$q_1$} &\qw& \ctrl{4} & \qw & \qw & \qw& \qw& \qw& \qw & \qw\\
\lstick{$q_2$} &\qw&\qw& \ctrl{4} & \qw & \qw & \qw & \qw& \qw& \qw\\
\lstick{$q_3$} &\qw&\qw&\qw& \ctrl{4} & \qw & \qw & \qw & \qw& \qw\\
\lstick{$q_4$} & \gate{R_x(\pi/4)} &\qw&\qw&\qw& \gate{R_y(\pi/4)} & \ctrl{1}&\qw&\qw & \qw \\
\lstick{$q_5$} &\qw& \gate{R_x(\pi/4)} &\qw&\qw& \gate{R_y(\pi/4)} & \targ{}& \ctrl{1}& \qw& \qw  \\
\lstick{$q_6$} &\qw&\qw& \gate{R_x(\pi/4)} &\qw& \gate{R_y(\pi/4)} & \qw&\targ{}&\ctrl{1}& \qw  \\
\lstick{$q_7$} &\qw&\qw&\qw& \gate{R_x(\pi/4)} & \gate{R_y(\pi/4)} & \qw& \qw& \targ{} &\qw
\end{quantikz}}
\caption{Linear Entanglement Binding Circuit with 4 qubits.}
\label{fig: Linear Entanglement}
\end{figure}
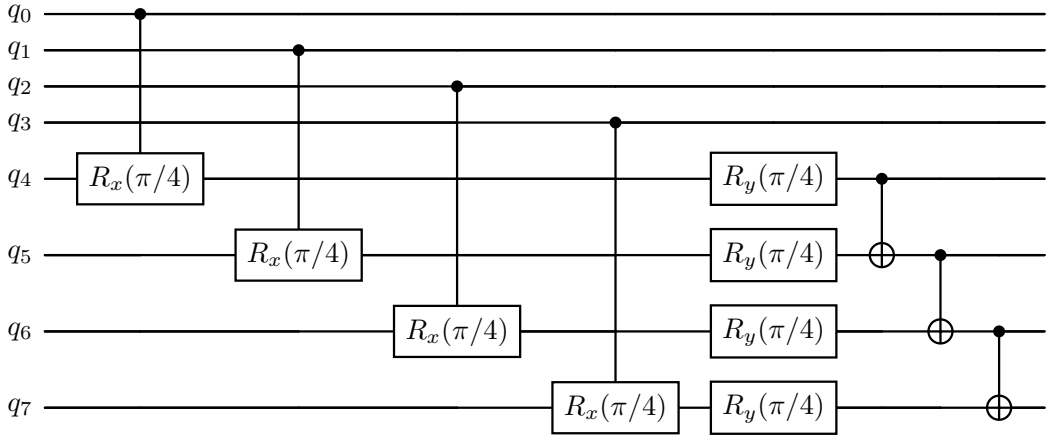
In this structure which, qubits in the second half (indexed from \( n \) to \( 2n-1 \)) are entangled sequentially in a linear (chain-like) configuration. Each qubit is entangled directly with its immediate neighbor, resulting in a simpler, lower-complexity structure.

The \textbf{unitary operator} representing this structure is expressed as follows:
\begin{equation}
U_{\text{linear}} = U_{\text{chain}} \cdot U_{\text{param}}
\end{equation}
where
\begin{equation}
U_{\text{chain}} = \prod_{i=n}^{2n-2} CX_{i,\, i+1},
\end{equation}
and
\begin{equation}
U_{\text{param}} = \prod_{i=0}^{n-1}\left[RY\left(\frac{\pi}{4}\right)_{n+i}\cdot CRX(\theta_i)_{i,\,n+i}\right].
\end{equation}

\subsection{Ring (Circular) Entanglement}
The circuit for the example with four qubits is shown in the figure.

\begin{figure}[ht]
\centering
\resizebox{\textwidth}{!}{\begin{quantikz}[row sep=0.3cm, column sep=0.4cm]
\lstick{$q_0$} & \ctrl{4} & \qw & \qw & \qw & \qw& \qw& \qw& \qw& \qw&\qw\\
\lstick{$q_1$} &\qw& \ctrl{4} & \qw & \qw & \qw& \qw& \qw& \qw & \qw&\qw\\
\lstick{$q_2$} &\qw&\qw& \ctrl{4} & \qw & \qw & \qw & \qw& \qw& \qw&\qw\\
\lstick{$q_3$} &\qw&\qw&\qw& \ctrl{4} & \qw & \qw & \qw & \qw& \qw&\qw\\
\lstick{$q_4$} & \gate{R_x(\pi/4)} &\qw&\qw&\qw& \gate{R_y(\pi/4)} & \ctrl{1}&\qw&\qw &\targ{}& \qw \\
\lstick{$q_5$} &\qw& \gate{R_x(\pi/4)} &\qw&\qw& \gate{R_y(\pi/4)} & \targ{}& \ctrl{1}& \qw& \qw &\qw \\
\lstick{$q_6$} &\qw&\qw& \gate{R_x(\pi/4)} &\qw& \gate{R_y(\pi/4)} & \qw&\targ{}&\ctrl{1}& \qw &\qw \\
\lstick{$q_7$} &\qw&\qw&\qw& \gate{R_x(\pi/4)} & \gate{R_y(\pi/4)} & \qw& \qw& \targ{} &\ctrl{-3}&\qw
\end{quantikz}}
\caption{Ring Entanglement Binding Circuit with 4 qubits.}
\label{fig: Ring Entanglement}
\end{figure}
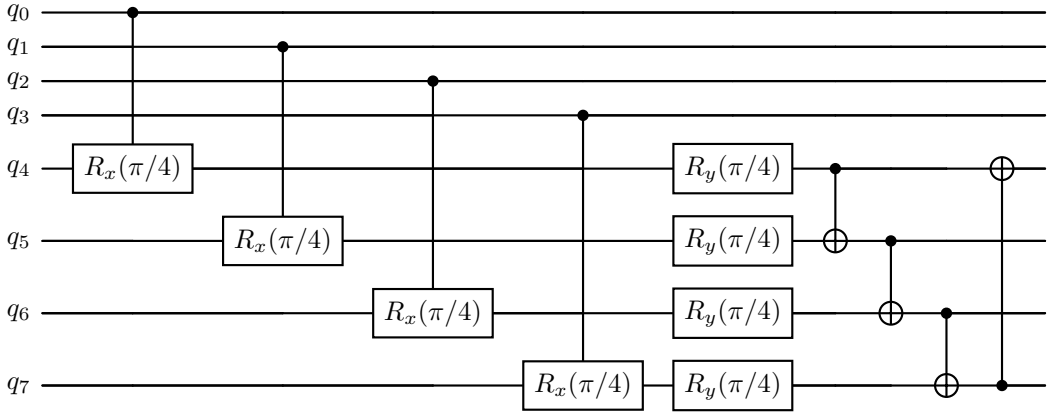

In this entanglement structure, the qubits in the latter half form a closed loop. Each qubit is entangled directly with its neighbor, including an additional connection between the first and last qubits to complete the ring. This structure is slightly more complex than linear entanglement but simpler than the fully connected entanglement. 

The \textbf{unitary operator} for ring entanglement is expressed as:
\begin{equation}
U_{\text{ring}} = U_{\text{circle}} \cdot U_{\text{param}}
\end{equation}
where
\begin{equation}
U_{\text{circle}} = \prod_{i=n}^{2n-1} CX_{i,\,((i-n+1)\mod n)+n},
\end{equation}
and \(U_{\text{param}}\) is as defined above.

Here, qubits indexed \(n\) to \(2n-1\) form a circular chain, ensuring each qubit is entangled with exactly two neighbors.

\section{Complete Results for Classification}
\label{appendix:results}
In addition to the two-class evaluations presented in the main text, we extend our experiments to multi-class classification tasks with increasing numbers of classes. Specifically, we conduct classification from 3 to 10 classes on the ISOLET and MNIST datasets, and from 3 to 6 classes on the UCI HAR dataset.

These experiments follow the same QeHDC framework, using one pass training and fidelity-based inference under fixed qubit and dimension configurations. Class-balanced subsets are selected for each setting.

The goal is to evaluate the scalability and stability of our method under increasing classification complexity. The results, summarized in the Table \ref{tab:3classes} to \ref{tab:10classes}, show that although accuracy naturally decreases as the number of classes increases, our method maintains strong relative performance compared to classical HDC baselines, especially under low-dimensional constraints.

\section{Tomography Circuits and Resconstructed State Comparison}
\label{appendix:tomography}
This appendix illustrates the tomography process used in our quantum classification framework, including representative tomography circuits and a comparison between the ideal post-binding state and the reconstructed density matrix.

\begin{figure*}[htbp]
\centering
\resizebox{\textwidth}{!}{\includegraphics[scale = 0.35]{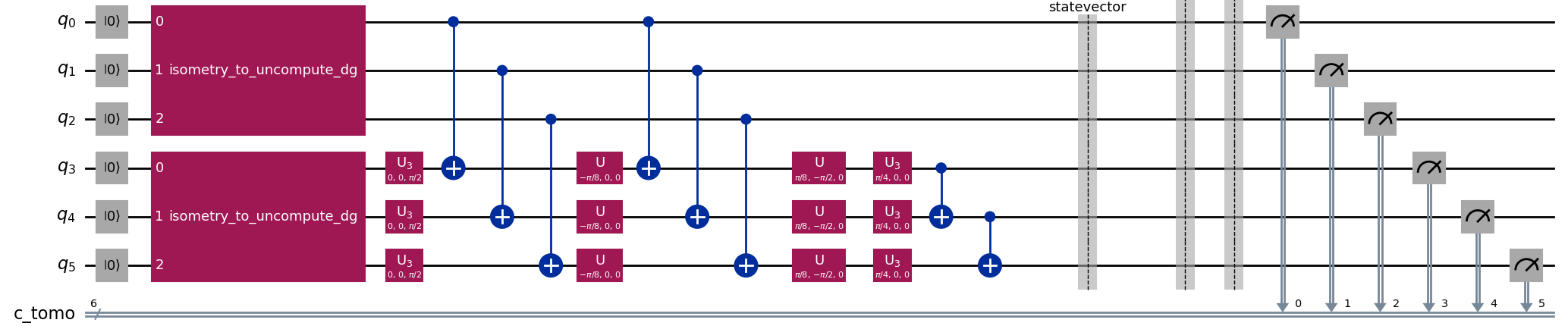}}
\caption{Quantum State Tomography Circuit (Z bias)}
\label{fig:tomography0}
\end{figure*}

\begin{figure*}[htbp]
\centering
\resizebox{\textwidth}{!}{
\includegraphics[scale = 0.32]{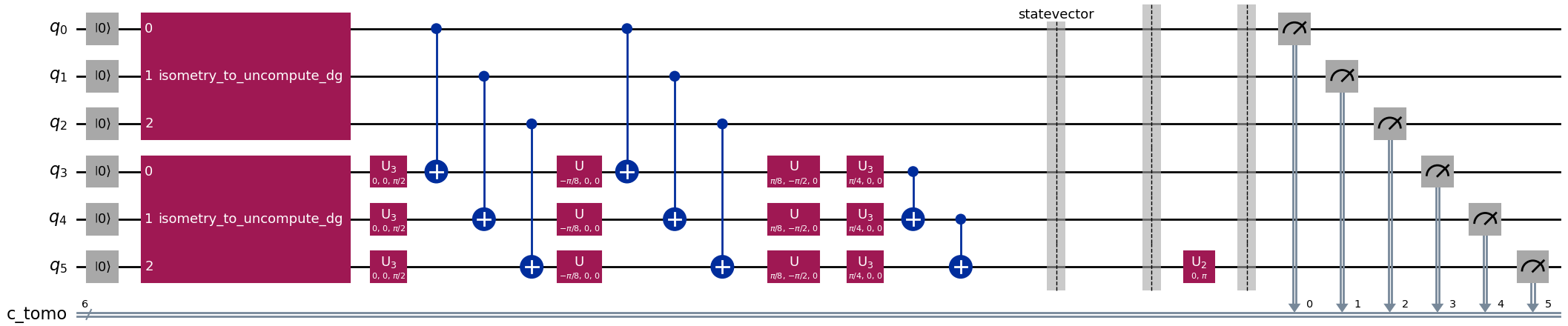}}
\caption{Quantum State Tomography Circuit (X bias)}
\label{fig:tomography1}
\end{figure*}

\begin{figure*}[htbp]
\centering
\resizebox{\textwidth}{!}{
\includegraphics[scale = 0.32]{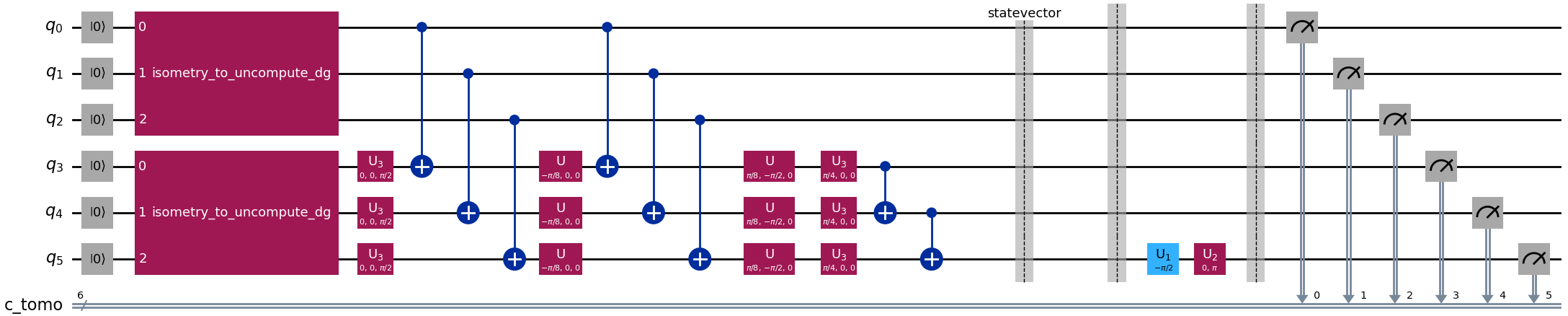}}
\caption{Quantum State Tomography Circuit (Y bias)}
\label{fig:tomography2}
\end{figure*}

Fig. \ref{fig:tomography0}, Fig. \ref{fig:tomography1}, and Fig. \ref{fig:tomography2}  show three example circuits generated by the tomography module. These circuits correspond to different measurement bases required for full quantum state reconstruction. To emphasize the core structure, all circuits are presented in a cleaned form with the initialization components removed.

\begin{figure*}[htbp]
\centering
\includegraphics[scale = 0.35]{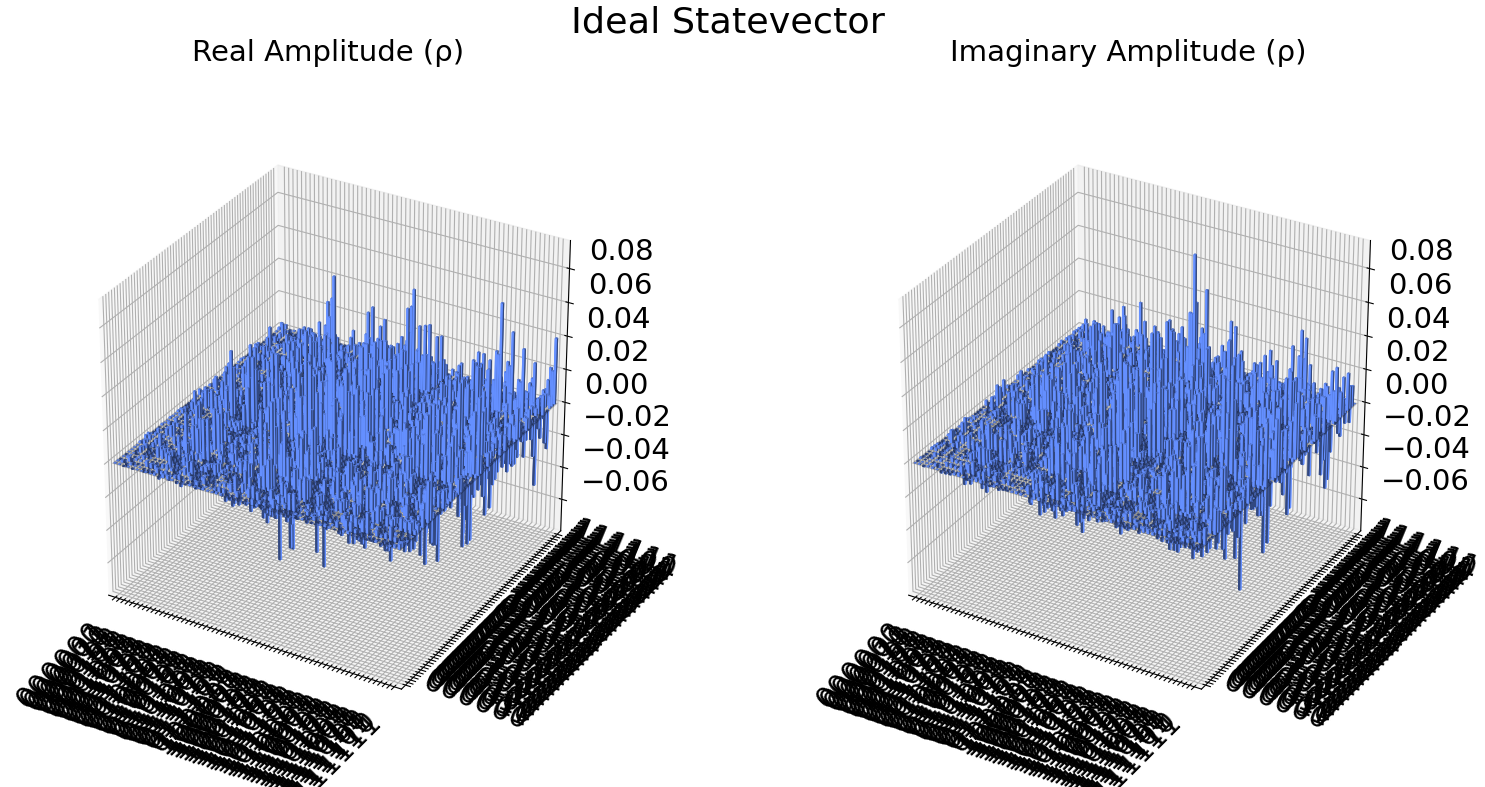}
\caption{Ideal Statevector (real and d Imaginary Components)}
\label{fig:IdealState}
\end{figure*}

\begin{figure*}[htbp]
\centering
\includegraphics[scale = 0.35]{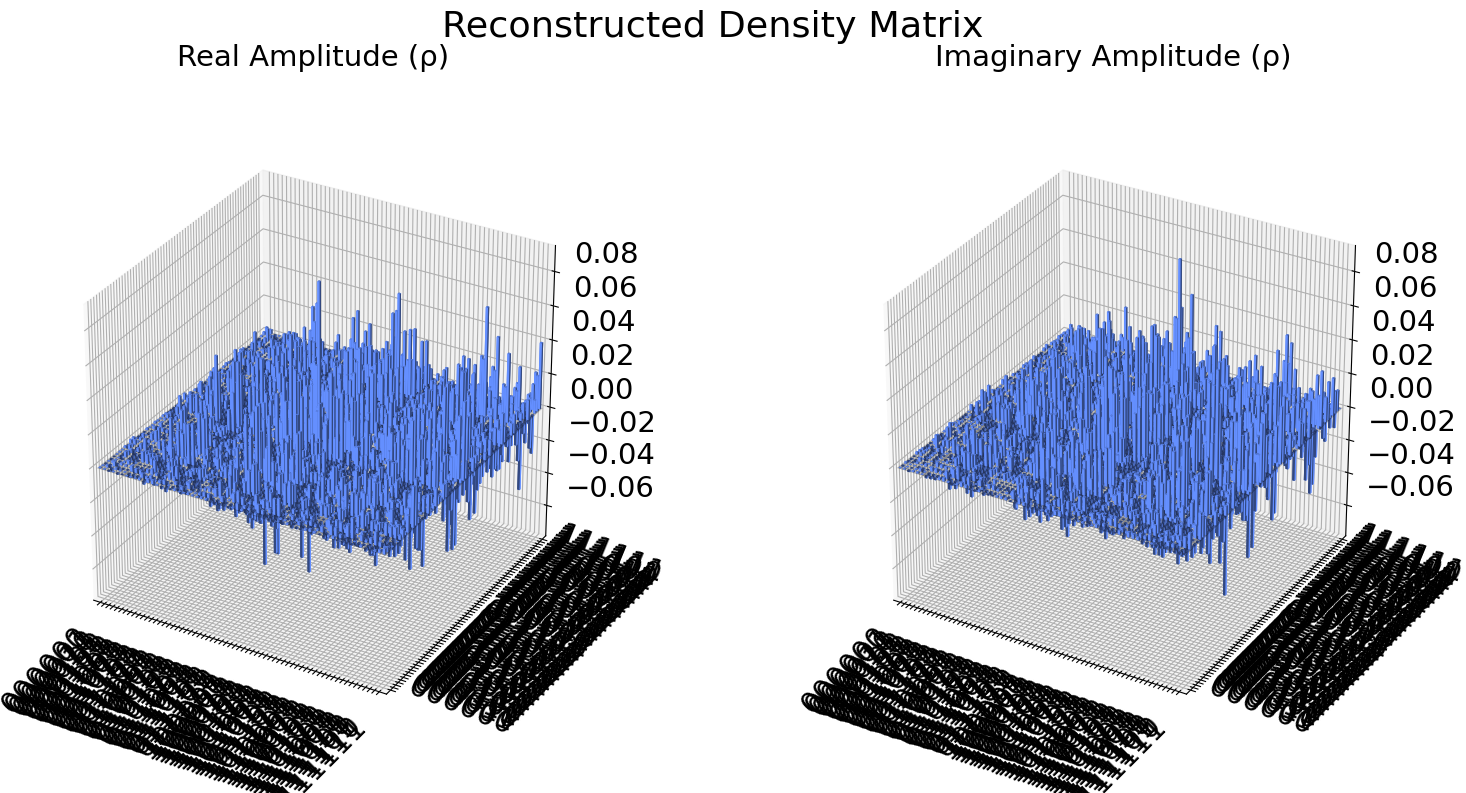}
\caption{Reconstructed Density Matrix (Real and Imaginary Components)}
\label{fig:ReconState}
\end{figure*}

After executing all tomography circuits, we reconstruct the output density matrix and compare it with the ideal quantum statevector obtained from the Aer simulator. Fig. \ref{fig:IdealState} and \ref{fig:ReconState} show the 3D cityscape visualizations of the complex amplitudes. The real and imaginary parts of both the ideal and reconstructed states are shown side by side for comparison.

As seen in the figures, the overall structure and amplitude distribution between the ideal and reconstructed states are highly consistent, validating the effectiveness of our QST pipeline even under limited measurement noise. These results demonstrate that the reconstructed quantum states retain sufficient fidelity for classification, as further supported by the numerical fidelity scores in the main text.

\begin{table*}[htbp]
\centering
\caption{3-Class Results (ISOLET, MNIST, UCI HAR)}
\label{tab:3classes}
\resizebox{\textwidth}{!}{
\begin{tabular}{l|ccc|ccc|ccc}
\toprule
\multirow{2}{*}{Method} & \multicolumn{3}{c|}{ISOLET-3Classes} & \multicolumn{3}{c|}{MNIST-3Classes} & \multicolumn{3}{c}{UCI HAR-3Classes} \\
\cmidrule(lr){2-4} \cmidrule(lr){5-7} \cmidrule(lr){8-10}
                        & 16D & 32D & 64D & 16D & 32D & 64D & 16D & 32D & 64D \\
\midrule
Vanilla   & 56.85$\pm$6.04 & 65.37$\pm$2.73 & 71.67$\pm$5.90 & 54.90$\pm$7.66 & 76.18$\pm$4.68 & 83.34$\pm$2.26 & 47.85$\pm$2.16 & 54.29$\pm$1.29 & 54.63$\pm$1.98 \\
AdaptHD   & 54.81$\pm$10.82 & 63.89$\pm$7.26 & 75.37$\pm$4.48 & 55.36$\pm$11.27 & 71.80$\pm$6.06 & 83.53$\pm$1.92 & 45.93$\pm$5.30 & 46.84$\pm$4.49 & 55.06$\pm$4.18 \\
OnlineHD  & 86.85$\pm$5.45  & 94.81$\pm$1.89 & 97.41$\pm$1.39 & 74.96$\pm$8.72  & 83.15$\pm$2.88 & 94.53$\pm$0.41 & 65.66$\pm$7.46 & 77.34$\pm$5.70 & 83.75$\pm$2.21 \\
NeuralHD  & 86.48$\pm$2.28  & 93.89$\pm$0.79 & 97.59$\pm$0.94 & 79.84$\pm$3.33  & 88.20$\pm$2.15 & 93.37$\pm$1.14 & 68.42$\pm$3.57 & 77.72$\pm$0.51 & 84.67$\pm$1.83 \\

CompHD    & 52.78$\pm$7.74  & 63.33$\pm$2.36 & 68.15$\pm$4.12 & 49.66$\pm$3.36  & 59.93$\pm$1.56 & 75.29$\pm$1.62 & 45.06$\pm$1.18 & 46.07$\pm$1.85 & 53.06$\pm$3.37 \\
SparseHD  & 51.48$\pm$5.00  & 61.48$\pm$7.61 & 73.33$\pm$7.86 & 44.46$\pm$12.26 & 43.00$\pm$4.77 & 64.65$\pm$17.43 & 41.14$\pm$3.05 & 46.43$\pm$3.78 & 51.43$\pm$3.60 \\
QuantHD   & 52.96$\pm$8.95  & 58.89$\pm$3.63 & 71.30$\pm$2.14 & 64.07$\pm$5.01  & 67.10$\pm$5.01 & 67.92$\pm$1.87 & 38.02$\pm$8.45 & 44.97$\pm$3.41 & 50.54$\pm$5.82 \\
LeHDC     & 46.67$\pm$7.07  & 48.15$\pm$2.73 & 53.89$\pm$9.50 & 44.55$\pm$5.35  & 53.21$\pm$5.02 & 57.91$\pm$11.07 & 37.66$\pm$0.65 & 38.62$\pm$2.94 & 42.78$\pm$2.03 \\

\bottomrule
\end{tabular}}
\end{table*}

\begin{table*}[htbp]
\centering
\caption{4-Class Results (ISOLET, MNIST, UCI HAR)}
\label{tab:4classes}
\resizebox{\textwidth}{!}{
\begin{tabular}{l|ccc|ccc|ccc}\toprule
\multirow{2}{*}{Method} & \multicolumn{3}{c|}{ISOLET-4Classes} & \multicolumn{3}{c|}{MNIST-4Classes} & \multicolumn{3}{c}{UCI HAR-4Classes} \\
\cmidrule(lr){2-4} \cmidrule(lr){5-7} \cmidrule(lr){8-10}
                        & 16D & 32D & 64D & 16D & 32D & 64D & 16D & 32D & 64D \\
\midrule
Vanilla   & 43.75$\pm$6.49  & 59.03$\pm$6.33  & 65.83$\pm$0.59  & 47.17$\pm$2.89  & 62.33$\pm$3.42  & 75.81$\pm$2.35  & 55.75$\pm$4.70  & 62.74$\pm$2.83  & 69.36$\pm$1.56 \\
AdaptHD   & 33.61$\pm$6.13  & 45.14$\pm$6.58  & 58.06$\pm$1.37  & 34.86$\pm$7.01  & 62.61$\pm$6.05  & 71.77$\pm$5.90  & 46.96$\pm$9.54  & 51.24$\pm$8.91  & 65.87$\pm$5.50 \\
OnlineHD  & 79.44$\pm$2.60  & 84.17$\pm$1.89  & 94.58$\pm$0.68  & 61.32$\pm$3.67  & 78.00$\pm$2.44  & 87.53$\pm$2.04  & 72.10$\pm$2.94  & 81.82$\pm$2.17  & 89.24$\pm$1.55 \\
NeuralHD  & 80.69$\pm$1.87  & 89.17$\pm$2.91  & 95.28$\pm$1.04  & 73.07$\pm$4.01  & 80.76$\pm$2.49  & 87.82$\pm$0.62  & 77.53$\pm$3.43  & 83.28$\pm$1.20  & 86.88$\pm$1.54 \\

CompHD    & 39.44$\pm$0.71  & 46.53$\pm$2.60  & 59.72$\pm$5.40  & 37.90$\pm$2.92  & 52.08$\pm$1.22  & 64.29$\pm$1.58  & 42.55$\pm$11.07 & 56.59$\pm$1.59  & 61.87$\pm$5.04 \\
SparseHD  & 42.08$\pm$3.25  & 41.94$\pm$4.82  & 61.67$\pm$5.31  & 35.54$\pm$4.85  & 37.40$\pm$4.10  & 38.35$\pm$10.15 & 39.94$\pm$3.58  & 50.66$\pm$2.50  & 56.37$\pm$4.04 \\
QuantHD   & 34.17$\pm$5.62  & 52.50$\pm$3.25  & 55.00$\pm$2.91  & 43.78$\pm$0.69  & 43.47$\pm$1.50  & 50.85$\pm$7.99  & 32.14$\pm$6.62  & 52.48$\pm$2.35  & 60.21$\pm$6.60 \\
LeHDC     & 32.64$\pm$3.99  & 40.83$\pm$2.57  & 38.33$\pm$7.27  & 32.79$\pm$5.02  & 38.92$\pm$8.12  & 50.01$\pm$5.80  & 43.91$\pm$5.97  & 40.47$\pm$3.11  & 47.46$\pm$1.29 \\

\bottomrule
\end{tabular}}
\end{table*}

\begin{table*}[htbp]
\centering
\caption{5-Class Results (ISOLET, MNIST, UCI HAR)}
\label{tab:5classes}
\resizebox{\textwidth}{!}{
\begin{tabular}{l|ccc|ccc|ccc}
\toprule
\multirow{2}{*}{Method} & \multicolumn{3}{c|}{ISOLET-5Classes} & \multicolumn{3}{c|}{MNIST-5Classes} & \multicolumn{3}{c}{UCI HAR-5Classes} \\
\cmidrule(lr){2-4} \cmidrule(lr){5-7} \cmidrule(lr){8-10}
                        & 16D & 32D & 64D & 16D & 32D & 64D & 16D & 32D & 64D \\
\midrule
Vanilla   & 34.33$\pm$2.13  & 54.89$\pm$4.12  & 56.11$\pm$5.65  & 44.54$\pm$4.14  & 58.66$\pm$3.18  & 71.17$\pm$3.59  & 47.25$\pm$3.31  & 51.81$\pm$2.70  & 59.97$\pm$3.75 \\
AdaptHD   & 32.56$\pm$4.73  & 42.56$\pm$10.02 & 48.11$\pm$4.94  & 34.43$\pm$5.58  & 46.27$\pm$5.24  & 62.39$\pm$11.57 & 32.45$\pm$4.27  & 49.07$\pm$2.78  & 54.23$\pm$4.15 \\
OnlineHD  & 62.22$\pm$3.45  & 78.44$\pm$0.79  & 87.00$\pm$1.25  & 53.30$\pm$10.33 & 76.50$\pm$2.72  & 86.98$\pm$0.93  & 62.14$\pm$4.12  & 75.19$\pm$3.11  & 84.67$\pm$1.73 \\
NeuralHD  & 70.78$\pm$4.36  & 80.33$\pm$1.96  & 88.78$\pm$0.87  & 69.59$\pm$0.80  & 79.67$\pm$3.36  & 86.95$\pm$1.38  & 71.44$\pm$3.02  & 79.52$\pm$1.56  & 82.85$\pm$0.54 \\

CompHD    & 31.22$\pm$4.01  & 36.89$\pm$4.31  & 46.67$\pm$0.72  & 38.79$\pm$6.67  & 48.76$\pm$3.28  & 57.24$\pm$0.80  & 41.96$\pm$4.42  & 45.39$\pm$0.50  & 54.54$\pm$2.33 \\
SparseHD  & 23.78$\pm$3.50  & 38.78$\pm$0.57  & 44.78$\pm$2.78  & 28.71$\pm$2.32  & 30.85$\pm$3.36  & 35.12$\pm$3.32  & 33.87$\pm$9.18  & 38.87$\pm$4.20  & 47.40$\pm$0.76 \\
QuantHD   & 24.22$\pm$5.97  & 34.67$\pm$4.23  & 50.11$\pm$5.12  & 29.78$\pm$8.10  & 36.56$\pm$6.67  & 53.29$\pm$1.56  & 34.34$\pm$1.32  & 43.62$\pm$5.63  & 44.48$\pm$3.34 \\
LeHDC     & 25.67$\pm$2.42  & 31.33$\pm$2.33  & 32.56$\pm$4.24  & 32.54$\pm$3.07  & 30.31$\pm$3.60  & 30.73$\pm$5.82  & 27.62$\pm$5.46  & 33.89$\pm$4.39  & 38.38$\pm$1.14 \\

\bottomrule
\end{tabular}}
\end{table*}

\begin{table*}[htbp]
\centering
\caption{6-Class Results (ISOLET, MNIST, UCI HAR)}
\label{tab:6classes}
\resizebox{\textwidth}{!}{
\begin{tabular}{l|ccc|ccc|ccc}
\toprule
\multirow{2}{*}{Method} & \multicolumn{3}{c|}{ISOLET-6Classes} & \multicolumn{3}{c|}{MNIST-6Classes} & \multicolumn{3}{c}{UCI HAR-6Classes} \\
\cmidrule(lr){2-4} \cmidrule(lr){5-7} \cmidrule(lr){8-10}
                        & 16D & 32D & 64D & 16D & 32D & 64D & 16D & 32D & 64D \\
\midrule
Vanilla   & 38.70$\pm$3.46  & 53.06$\pm$3.15  & 55.46$\pm$2.93  & 44.58$\pm$3.79  & 47.49$\pm$5.53  & 57.42$\pm$2.79  & 36.91$\pm$4.12  & 46.08$\pm$5.89  & 57.26$\pm$0.48 \\
AdaptHD   & 30.56$\pm$1.64  & 39.17$\pm$7.20  & 54.26$\pm$0.13  & 27.00$\pm$9.06  & 32.54$\pm$8.22  & 55.82$\pm$6.17  & 38.91$\pm$10.41 & 42.11$\pm$2.05  & 51.89$\pm$4.34 \\
OnlineHD  & 63.52$\pm$3.10  & 80.93$\pm$3.67  & 88.24$\pm$1.82  & 51.21$\pm$3.14  & 67.49$\pm$1.42  & 80.39$\pm$1.85  & 68.47$\pm$2.94  & 76.96$\pm$3.53  & 85.11$\pm$0.67 \\
NeuralHD  & 75.09$\pm$0.92  & 82.69$\pm$2.16  & 87.59$\pm$1.33  & 63.05$\pm$1.87  & 71.76$\pm$0.66  & 82.70$\pm$0.98  & 73.56$\pm$1.65  & 80.24$\pm$0.98  & 84.11$\pm$0.99 \\

CompHD    & 29.44$\pm$4.47  & 41.67$\pm$3.20  & 52.50$\pm$2.40  & 30.46$\pm$1.68  & 40.80$\pm$5.94  & 51.21$\pm$4.03  & 32.47$\pm$4.98  & 41.42$\pm$7.17  & 51.95$\pm$3.16 \\
SparseHD  & 23.98$\pm$4.81  & 36.76$\pm$2.92  & 50.74$\pm$1.98  & 23.76$\pm$4.26  & 24.99$\pm$5.00  & 27.75$\pm$5.64  & 30.84$\pm$1.15  & 35.69$\pm$3.96  & 41.78$\pm$4.06 \\
QuantHD   & 29.81$\pm$3.19  & 40.19$\pm$2.87  & 46.39$\pm$4.74  & 31.76$\pm$10.03 & 36.68$\pm$3.68  & 39.92$\pm$6.45  & 28.06$\pm$2.35  & 40.48$\pm$3.19  & 38.89$\pm$0.99 \\
LeHDC     & 23.61$\pm$3.60  & 28.15$\pm$4.93  & 34.35$\pm$4.67  & 25.86$\pm$4.28  & 30.90$\pm$2.35  & 32.76$\pm$4.65  & 21.67$\pm$2.43  & 31.21$\pm$3.91  & 34.02$\pm$6.37 \\
\bottomrule
\end{tabular}}
\end{table*}

\begin{table*}[htbp]
\centering
\caption{7-Class Results (ISOLET and MNIST)}
\label{tab:7classes}
\resizebox{\textwidth}{!}{
\begin{tabular}{l|ccc|ccc}
\toprule
\multirow{2}{*}{Method} & \multicolumn{3}{c|}{ISOLET-7Classes} & \multicolumn{3}{c}{MNIST-7Classes} \\
\cmidrule(lr){2-4} \cmidrule(lr){5-7}
                        & 16D & 32D & 64D & 16D & 32D & 64D \\
\midrule
Vanilla   & 36.75$\pm$7.88 & 47.94$\pm$2.59 & 54.21$\pm$3.77 & 34.69$\pm$7.28 & 43.94$\pm$1.91 & 53.18$\pm$1.64 \\
AdaptHD   & 26.43$\pm$3.03 & 30.32$\pm$2.35 & 40.48$\pm$6.62 & 32.28$\pm$6.47 & 30.57$\pm$4.43 & 54.22$\pm$2.76 \\
OnlineHD  & 67.14$\pm$1.47 & 80.71$\pm$1.85 & 88.73$\pm$1.63 & 50.60$\pm$3.36 & 63.90$\pm$1.21 & 80.04$\pm$1.12 \\
NeuralHD  & 71.51$\pm$0.59 & 81.75$\pm$3.34 & 87.70$\pm$1.81 & 52.95$\pm$2.54 & 71.00$\pm$1.94 & 80.43$\pm$1.30 \\

CompHD    & 28.49$\pm$4.89 & 30.32$\pm$0.74 & 45.95$\pm$2.21 & 26.21$\pm$2.02 & 36.59$\pm$0.51 & 45.29$\pm$2.88 \\
SparseHD  & 24.92$\pm$2.65 & 31.35$\pm$3.57 & 41.59$\pm$2.63 & 22.16$\pm$2.90 & 19.84$\pm$2.22 & 25.70$\pm$4.52 \\
QuantHD   & 30.16$\pm$5.63 & 37.54$\pm$5.35 & 44.05$\pm$4.60 & 25.84$\pm$4.11 & 31.91$\pm$6.93 & 39.29$\pm$6.00 \\
LeHDC     & 20.48$\pm$3.41 & 22.14$\pm$2.44 & 29.44$\pm$1.29 & 22.17$\pm$4.49 & 24.20$\pm$2.38 & 34.02$\pm$6.17 \\

\bottomrule
\end{tabular}}
\end{table*}

\begin{table*}[htbp]
\centering
\caption{8-Class Results (ISOLET and MNIST)}
\label{tab:8classes}
\resizebox{\textwidth}{!}{
\begin{tabular}{l|ccc|ccc}
\toprule
\multirow{2}{*}{Method} & \multicolumn{3}{c|}{ISOLET-8Classes} & \multicolumn{3}{c}{MNIST-8Classes} \\
\cmidrule(lr){2-4} \cmidrule(lr){5-7}
                        & 16D & 32D & 64D & 16D & 32D & 64D \\
\midrule
Vanilla   & 33.33$\pm$3.96 & 40.49$\pm$3.45 & 57.85$\pm$3.47 & 30.77$\pm$1.49 & 43.19$\pm$3.33 & 58.51$\pm$1.72 \\
AdaptHD   & 27.50$\pm$1.53 & 37.71$\pm$6.78 & 46.67$\pm$7.04 & 24.16$\pm$1.67 & 33.92$\pm$5.43 & 47.44$\pm$7.57 \\
OnlineHD  & 65.42$\pm$5.35 & 80.90$\pm$1.37 & 88.47$\pm$0.39 & 42.01$\pm$3.21 & 60.49$\pm$1.76 & 77.44$\pm$1.54 \\
NeuralHD  & 70.90$\pm$1.87 & 85.76$\pm$2.06 & 91.39$\pm$1.43 & 56.90$\pm$2.44 & 69.39$\pm$2.67 & 80.10$\pm$0.66 \\

CompHD    & 25.35$\pm$0.55 & 32.85$\pm$0.94 & 42.57$\pm$4.53 & 24.54$\pm$3.71 & 32.65$\pm$2.69 & 42.71$\pm$0.97 \\
SparseHD  & 17.50$\pm$4.36 & 30.00$\pm$4.34 & 35.14$\pm$3.09 & 22.64$\pm$5.33 & 19.76$\pm$4.34 & 26.02$\pm$1.65 \\
QuantHD   & 18.61$\pm$4.03 & 30.62$\pm$2.86 & 39.79$\pm$1.84 & 16.57$\pm$6.14 & 37.69$\pm$0.89 & 39.40$\pm$2.17 \\
LeHDC     & 16.32$\pm$1.90 & 19.93$\pm$1.82 & 28.19$\pm$2.77 & 16.70$\pm$4.37 & 23.32$\pm$1.14 & 29.05$\pm$3.52 \\

\bottomrule
\end{tabular}}
\end{table*}

\begin{table*}[htbp]
\centering
\caption{9-Class Results (ISOLET and MNIST)}
\label{tab:9classes}
\resizebox{\textwidth}{!}{
\begin{tabular}{l|ccc|ccc}
\toprule
\multirow{2}{*}{Method} & \multicolumn{3}{c|}{ISOLET-9Classes} & \multicolumn{3}{c}{MNIST-9Classes} \\
\cmidrule(lr){2-4} \cmidrule(lr){5-7}
                        & 16D & 32D & 64D & 16D & 32D & 64D \\
\midrule
Vanilla   & 29.44$\pm$4.92 & 41.98$\pm$1.83 & 57.16$\pm$2.60 & 27.87$\pm$0.53 & 39.08$\pm$4.12 & 52.22$\pm$2.58 \\
AdaptHD   & 25.43$\pm$2.57 & 24.01$\pm$6.12 & 47.04$\pm$6.91 & 21.74$\pm$3.17 & 23.52$\pm$7.59 & 47.90$\pm$3.54 \\
OnlineHD  & 62.78$\pm$4.74 & 84.51$\pm$0.76 & 88.15$\pm$1.46 & 37.52$\pm$4.24 & 54.04$\pm$2.83 & 71.56$\pm$1.28 \\
NeuralHD  & 70.62$\pm$3.33 & 83.21$\pm$2.06 & 90.49$\pm$1.01 & 53.65$\pm$1.53 & 68.81$\pm$1.62 & 77.13$\pm$0.85 \\
CompHD    & 26.91$\pm$1.44 & 30.43$\pm$1.57 & 41.54$\pm$5.18 & 23.36$\pm$1.98 & 29.93$\pm$2.70 & 38.34$\pm$1.48 \\
SparseHD  & 17.35$\pm$4.61 & 24.26$\pm$0.69 & 37.41$\pm$4.45 & 15.79$\pm$2.40 & 18.57$\pm$2.11 & 21.57$\pm$3.18 \\
QuantHD   & 20.37$\pm$2.72 & 26.42$\pm$3.29 & 38.02$\pm$3.97 & 20.80$\pm$2.57 & 29.95$\pm$0.28 & 36.52$\pm$4.86 \\
LeHDC     & 20.06$\pm$4.52 & 22.65$\pm$4.83 & 31.60$\pm$1.29 & 19.33$\pm$0.34 & 18.30$\pm$0.39 & 25.10$\pm$4.19 \\
\bottomrule
\end{tabular}}
\end{table*}

\begin{table*}[htbp]
\centering
\caption{10-Class Results (ISOLET and MNIST)}
\label{tab:10classes}
\resizebox{\textwidth}{!}{
\begin{tabular}{l|ccc|ccc}
\toprule
\multirow{2}{*}{Method} & \multicolumn{3}{c|}{ISOLET-10Classes} & \multicolumn{3}{c}{MNIST-10Classes} \\
\cmidrule(lr){2-4} \cmidrule(lr){5-7}
                        & 16D & 32D & 64D & 16D & 32D & 64D \\
\midrule
Vanilla   & 27.72$\pm$2.73 & 39.78$\pm$3.68 & 54.11$\pm$2.70 & 23.49$\pm$3.11 & 30.96$\pm$4.59 & 46.61$\pm$4.41 \\
AdaptHD   & 16.28$\pm$1.95 & 25.11$\pm$2.91 & 47.33$\pm$1.78 & 16.42$\pm$4.03 & 21.82$\pm$4.04 & 36.27$\pm$4.52 \\
OnlineHD  & 62.89$\pm$4.83 & 81.89$\pm$1.47 & 90.67$\pm$1.30 & 34.03$\pm$1.42 & 53.24$\pm$4.90 & 69.79$\pm$0.30 \\
NeuralHD  & 68.67$\pm$3.42 & 85.61$\pm$1.80 & 90.83$\pm$0.49 & 47.04$\pm$1.52 & 63.92$\pm$1.89 & 75.73$\pm$0.36 \\
DistHD    & 75.67$\pm$3.31 & 86.00$\pm$2.12 & 92.94$\pm$0.63 & 50.84$\pm$2.48 & 67.90$\pm$2.13 & 78.97$\pm$1.54 \\
CompHD    & 21.39$\pm$4.47 & 32.44$\pm$3.85 & 35.22$\pm$1.61 & 20.25$\pm$0.67 & 26.88$\pm$1.51 & 36.88$\pm$2.37 \\
SparseHD  & 13.94$\pm$1.59 & 22.06$\pm$2.79 & 36.11$\pm$3.94 & 14.07$\pm$1.66 & 14.53$\pm$0.92 & 23.50$\pm$3.23 \\
QuantHD   & 21.61$\pm$2.09 & 28.33$\pm$4.90 & 39.39$\pm$6.28 & 21.17$\pm$2.65 & 25.27$\pm$1.64 & 30.93$\pm$2.74 \\
LeHDC     & 14.78$\pm$5.18 & 17.17$\pm$4.25 & 22.33$\pm$1.91 & 15.43$\pm$3.16 & 17.43$\pm$4.05 & 26.50$\pm$2.60 \\

\bottomrule
\end{tabular}}
\end{table*}

\clearpage

\end{document}